# Graph Theory Applications in Advanced Geospatial Research

Surajit Ghosh[1,*], Archita Mallick[2], Anuva Chowdhury[3], Kounik De Sarkar[4]

[1]International Water Management Institute, Colombo, Sri Lanka
[2]Techno India University, Kolkata, West Bengal, India
[3]Birla Institute of Technology Mesra, Jharkhand, India
[4]National Institute of Technology Durgapur, West Bengal, India

*Contact: s.ghosh@cgiar.org*

**Abstract**
Geospatial sciences include a wide range of applications, from environmental monitoring transportation to infrastructure planning, as well as location-based analysis and services. Graph theory algorithms in mathematics have emerged as indispensable tools in these domains due to their capability to model and analyse spatial relationships efficiently. This article explores the applications of graph theory algorithms in geospatial sciences, highlighting their role in network analysis, spatial connectivity, geographic information systems, and various other spatial problem-solving scenarios like digital twin. The article provides a comprehensive idea about graph theory's key concepts and algorithms that assist the geospatial modelling processes and insights into real-world geospatial challenges and opportunities. It lists the extensive research, innovative technologies and methodologies implemented in this domain.

**Keywords**: Graph theory, Geospatial Science, Digital twin

## 1. Introduction

Geospatial science has developed as a vibrant field characterised by intellectual vigour, conceptual expansion, and improved analytical skills as a consequence of the Quantitative Revolution in the subject of geography through a spatially integrated socio-environmental science that outshines prior disciplinary ties, borders, and limitations (**Berry et al., 2008**). Geospatial science, commonly referred to as geomatics (**Aina 2012**), is a multidisciplinary discipline that focuses on comprehending, analysing, and visualising spatial data about the Earth's surface using information technology to describe the connections between geography, individuals, places, and Earth processes. Technologies like Global Positioning System (GPS), Geographic Information Systems (GIS), and remote sensing are frequently used as observational, measuring, and analytical tools, helping in the understanding of numerous events by providing the information with a spatial context. Geospatial technology is being used increasingly in every industry today, including resource management, disaster management, forestry, logistics, infrastructure planning, and the study of climate change and other environmental issues (**Dangermond and Goodchild, 2020**). Geospatial technology and the information created are becoming increasingly significant in all economic sectors, making the economy, society, and the environment an indispensable pillar of sustainable development. (**Scott and Rajabifard, 2017**). Thus, geospatial science and technology support disaster management (**Ghosh and Mukherjee, 2023**), infrastructure planning, environmental monitoring (**Sanyal and Chowdhury, 2023**), as well as location-based services.

Digital Twin's concept involves creating virtual models of physical objects and systems to reproduce their real-world counterparts as accurately as possible. They capture both the static and dynamic behaviour of objects or systems. Digital twin finds their applications in various fields, starting from real-time monitoring of objects, maintenance and optimisation of systems, designing prototypes virtually before building them to simulating and predicting climate change and monitoring the performance of aircraft and grid systems. In the backdrop of facing the change and modernisation of manufacturing sectors and changing to smart manufacturing, the digital twin, as a novel technological tool for implementing smart manufacturing, has drawn numerous scholars' research and discussion. Although this notion has been offered for some time, there are few project uses of a digital twin as a technical instrument, the referenceable expertise is barely any, and the reference content is the primarily theoretical and conceptual study (**Zhou et al., 2022**). Geospatial Digital Twin (GDT) emphasizes the geospatial attributes of the geographical settings, incorporating precise location and spatial layers for building a comprehensive knowledge of the spatial environment and its entities. Therefore, the geospatial concept and network between the entities are one of the core parts of GDT. Thus, implementing GDT is not a straightforward process and needs a variety of spatial computing technologies (**Eunus et al., 2023**), including graph theory.

Graph algorithms are used extensively in location-based services and analysis (**Wang et al., 2019**). Graph algorithms analyse spatial connections and relationships between two points (**Demšar et al. 2008**) or locations. Graph theory is the study of the mathematical structures known as graphs, which are used to represent pairwise interactions between objects (**Singh 2014**) (**Table 1**). Graphs may be spatial or non-spatial graphs, which further contain both directed and undirected graphs with weighted and unweighted components (**Table 2**). Real spatial or non-spatial networks possess characteristics specific to one of the following four graph types: regular, random, small world, and scale-free (**Anderson and Dragićević 2020**). An adjacency matrix can be used to depict the organisation of these graphs (**Anderson et al., 2020**).

The initial use of graph theory (**Euler et al., 1741**), 200 years earlier, was a location-based problem known as "Seven Bridges of Konigsberg", where Euler demonstrated that it was impossible to travel over all seven of the bridges that connect the islands without ever using the same bridge twice. This approach, also known as network science, is shown by the shortest path routing algorithms (**Table 3**) between two points (**Dijkstra et al., 1959**). The theorem was developed as a result of Euler's discoveries, which served as the cornerstone of network science. The findings also led to the conclusion that Graph Theory may be used to uncover and represent many structural properties (**Anderson et al., 2020**). In the 19$^{th}$ century, Cayley's studies formed the beginning of enumerative graph theory, using trees as the types of graphs, and focused on calculating the number of certain types of graphs (**Bell et al., 2015**). Social network analysis, one of the earliest fields of application (**de Nooy et al., 2005**), was where the three types of centrality metrics were initially established (**Freeman et al., 1979**). In social networks, vertices stand in for people or institutions, and edges show their connections to one another. In social network analysis, a person's reachability—or how readily information can go to that person—is described by their degree and closeness centralities. Journey from then till today, graph theory has extended its applications from social media network cybersecurity to fields of bioinformatics and cryptography. With the increase in data connectedness and breakthroughs in graph technology, valuable insights are obtained when integrated with queries, statistics, algorithms, ML, and AI (**Anderson et al., 2020**).

The objective of this article is to understand some fundamental concepts and examples of graph algorithms, their applications in geospatial science (**Figure 1** and **Table 4**), digital twin, and the methods by which geographic data, network sciences, and graph algorithms can be used to represent, analyse, and simulate complex geographical systems for better decision-making.

**2. Uses cases in Geospatial Research**

This section discusses advanced applications and research using graph algorithms. The below-mentioned applications vary over a range of domains like smart city planning, environment conservation, waste management, development of scientific tools and digital twin.

*2.1. Alternate route planning* (**Vanhove et al., 2010**)

The report uses the shortest path algorithms in a realistic situation where forbidden turns (Turn Prohibitions) and turn costs are taken into account. A solution to prohibited turns has been modelled by the Direct method, which is an adaptation of the Dijkstra algorithm. This method did not take into consideration Turn costs, so another method, Node Splitting, was proposed. Turn cost as well as turn prohibition restrictions were both given importance. A third method, the Line Graph method, was also proposed for the restrictions. This method is a graph transformation algorithm that has the freedom to use any shortest path algorithm. The second part of the report focused on alternative route planning for which the k shortest paths algorithm and a heuristic approach known as the deviation path algorithms (which follows Yen's approach) were used as a solution model. It has been observed that the heuristic method is always quicker than the exact approach and frequently misses none or very few pathways. The future scope suggests optimising the k shortest paths problem and the heuristic approach.

*2.2. Facility Search* (**Brost et al., 2014**)

The report presents an automated search method to find industrial facilities in remote sensing data. Graphs here encode properties and features of relationships and enable search methods that do not rely on specific predefined shapes. The methodology applies a unique structure of semantics graphs and efficient data processing technologies. It says that once the problem is reduced to finding patterns of nodes and edges in a graph, it becomes a problem of subgraph isomorphism. The research uses a methodology where images and tables from the data sources are added to a program in the graph construction phase, followed by a search graph program to find matches. Each land cover region had its node in the final graph, along with information like the kind of cover, the bounding box, the centroid, the area, and the eccentricity.

*2.3. Minimum Spanning Tool (MST) tool development* (**Dutta et al., 2014**)

The main topic of the report is the development of a GIS tool that employs Prim's Algorithm to generate the Minimum Spanning Tool (MST) of a road network while taking significant junction sites into account. Database creation and tool development are two parts that summarise the technique used for this project. The phase of tool development comes after database building, which comprises choosing the region that needs high-quality satellite imagery and digitising the photos. The outcomes from utilising this GIS tool demonstrate that there is better satisfaction when attempting to reduce the overall road length, which is a necessary condition for connecting important nodes (junctions) with one another.

*2.4. Forest patch connectivity diagnostics and prioritisation* (**Devi et al., 2013**)

Using graph theory, the report proposes a method for determining the ideal forest patch locations and threshold distances. Graph theory-based connectivity indices were used to establish the ideal threshold distance and its components. The connectivity investigated in this study is based on landscape structure, i.e., structural connectedness. It is used to evaluate graph theory applications and determine the best connectivity between forest landscapes. The status of fragmentation is determined by supplying patch counts and area range categories. The ArcGIS extension is used to obtain nodes and lengths concerning forest type polygons. The study emphasises how the use of graph theory made it easier to determine the best patches for connection and find suitable habitat patches for the preservation of biodiversity. The study demonstrated that hierarchical analysis of patch size, number, inter-distance, and relative importance was required to establish optimal connectivity.

*2.5. Network-based Geographic Automata* (**Anderson et al., 2020**)

The study provides details on network theory-related scientific research and assesses the likelihood of combining it with complex systems modelling techniques. It explores how static spatial network architectures change as a result of spatial dynamics and the inclusion of GAS modelling. The study provides a summary of descriptive graph theory measurement forms of theoretical graphs and characterises spatial social, ecological, and geophysical networks to assess their effectiveness. Current research trends include understanding the spatial dynamics of objects operating on spatial network structures, including network theory as GAS, and comprehending the intricately intertwined relationship between spatial structure and space-type dynamics.

*2.6. Modelling the effects of surface currents* (**Kasyk et al., 2016**)

This document outlines an approach to designing a graph representing the interactions between moving objects and surface currents. Multidimensional analysis is carried out by merging the graph with port charts. The ability to estimate the impacts of currents on moving dockside objects is made possible by the comprehensive knowledge of navigation and harbour infrastructure. In order to analyse how surface currents interact in a harbour's waters, a graph was modelled after processing and visualising multidimensional spatial data and after taking orthophotographs of the harbour in graphical format. The report provided techniques for graphically displaying crucial environmental data, and the outcomes were verified using hydrodynamic models.

*2.7. Landscape connectivity* (**Bunn et al., 2000**)

The study uses an examination of applying a graph-theoretical method to landscape connectivity using focal species. The technique of connectedness is used for two focus species. The MST serves as the graph's supporting structure, while the graph's diameter serves as a measure of the habitat mosaic's overall traversability. Distances are calculated as a series of least-cost paths on a cost surface that is suitable for the organism under study, and edges are determined by a dispersal probability matrix. There have been instances of graph operations like node removal and edge thresholding. It was discovered that the species' dispersal skills affected how the landscape was seen by each species. Additionally, it contributed to the discovery that one of the biggest dangers to loss of habitat in the world is habitat fragmentation. Identifying the patches where the study should be prioritised was also analysed.

*2.8. Rural electrification planning* (**Gadelha Filho et al., 2020**)

The research suggests a plan for rural electricity utilising topography analysis, graph theory, and GIS. Through an expansion of the MV distribution network, it seeks to design the electric network topology most economically, serving up to 100% of the local population. Three primary techniques are employed to determine the ideal electric network architecture: numerical (Mixed Integer Linear Programming or Non-Linear Programming), heuristic and metaheuristic, and graph theory. The medium and low voltage grid, which employs the minimal spanning tree and a branch exchange technique, is designed using the Reference Network model (REM). Graph theory is used to construct the grid routing algorithm. The results demonstrate that the total investment cost for line deployment was reduced by up to 47% using the innovative approach suggested (in comparison to a conventional minimal spanning tree procedure).

*2.9. Sustainable stormwater management* (**Kaur et al., 2022**)

The report outlines a method for identifying Blue Green Infrastructure (BGI) networks based on geospatial technologies. For effective stormwater management in metropolitan settings, graph theory is used. It also presents a reproducible approach by adding five important criteria to roads for suitability analysis. Determining a BGI network with a preference for stormwater management, detecting via satellite pictures patches of blue and green, and identifying prospective BGI corridors using graph theory were the three stages that were taken. A suitability analysis was carried out to choose the best location to develop the same. The study identifies places with a high potential for BGI network deployment and offers a simple idea of blue-green infrastructure above grey infrastructure for urban design and development that considers water concerns in any metropolis worldwide.

*2.10. Urban accessibility during an earthquake* (**Bono et al., 2011**)

In order to examine how accessibility declines when road networks are broken, the research presents an alternative way to visualise the accessibility landscape of cities following an earthquake. Because a line was split into two after disturbances, it was seen that the total number of segments making up the damaged network was higher than the original network. The difference in performance between the intact and damaged networks represents the possibility that the earthquake will cause a drop in transit efficiency. The study informed us of the challenges in allocating resources into a successful supply chain. An approach that combines spatial and network analysis was used to measure the degree of isolation of urban units as a result of substantial disturbances to the urban road network caused by collapsed buildings and debris. The method was found to be useful for displaying isolated urban areas and analysing accessibility from the road network. The preliminary analysis, which included graph deletion and erosion for all streets where interruptions were reported, yielded a more severe reduced accessibility result, and the current method proved to be more precise in determining the degree of isolation. The approach more accurately captures the impacts on geographically scattered but connected assets, which might be challenging to recognise. Making sense of the disruption that natural catastrophes produce, from a humanitarian standpoint and to the economic supply chain, may be done using even basic graph theoretic concepts combined with GIS.

*2.11. Emergency Evacuation Planning* (**Mohamed et al., 2018**)

According to the research, a precise and effective evacuation plan that includes numerous evacuation routes while promptly considering changing road conditions is critical to limiting damage. As a solution to this problem, shortest-path algorithms, decision-support systems, and GIS are being used. The proposed framework includes calculating the level of risk posed by the event (three emergency models are included), identifying all safe locations, calculating the radius of the affected area, and determining the best alternate routes for evacuation from within the buffer zone to safe destinations based on dynamic road conditions displayed on the map. A graph theory-based approach is proposed based on an established strategy for getting individuals to the nearest safe locations. It detects all secure venues and provides the best options. As a result, Dijkstra's algorithm is performed from one place within the buffer to every recognised safe destination, the shortest time to travel path and several alternate paths are estimated. A GIS-based evacuation planning (G-BEP) prototype is used to evaluate the suggested framework. The main contributions of this paper are the combination of (1) three different emergency danger-prediction models; (2) the development of the ArcGIS "Evacuation-Analysis Tool" toolbar; (3) the proposal of a framework that combines GIS capabilities, emergency simulations, shortest-paths algorithms, and factors in variables associated with dynamic road conditions to select the best alternate routes; and (4) identification of the nearest safe destinations outside the buffer zone.

*2.12. Vulnerability in Spatial Networks* (**Demšar et al., 2008**)

In order to identify crucial places in a spatial network, this article proposes a mathematical method for calculating the vulnerability of components of the network. The approach, which has been validated on the street network, incorporates dual graph modelling, connectivity analysis, and topological metrics. The method, which has its foundation in graph theory, examines the network's topology—that is, the connections between network segments—to pinpoint problematic components. For visualisation purposes, the geometry of the network's elements (*i.e.,* the precise physical placement of the streets) has been utilised, but the approach does not make use of any attribute data. As a result, the approach could be used with networks that do not have a fixed physical position but are thought of as spatial metaphors for non-geographic data. Although geographical positioning is the central idea and research area of GIS, it has recently become clear that long-established models and techniques used by geographers can be successfully used to represent things, phenomena, or methods with spatial features and behaviour.

*2.13. Flooding Similarity Mapping* (**Melnik et al., 2002**)

This paper introduces a general-purpose match algorithm built around a fixpoint computation. As its output, the algorithm generates a mapping between the corresponding nodes of two graphs (schemas, catalogues, or other data structures). Filters are used to choose a subset of the map based on the matching aim. Directed labelled graphs serve as the foundation of the inner data model that we employ for modelling and mapping. User research to assess the algorithm's performance for schema-matching tasks. The user research data was used to guide our evaluation and customisation of the algorithm for schema matching, using matching accuracy as the quality metric. The parameters of the algorithm and the filter that, on average, performed the best for all users and matching difficulties in our study were identified as a result of this examination. Overall labour reductions across all jobs for the average user were greater than 50%. The accuracy metric provided a pessimistic estimate; actual savings might have been considerably larger. A fixpoint formula with rapid convergence did not introduce accuracy penalties. The study says that the threshold filter worked the best. The inverse average provided the basis for the best formula for calculating propagation coefficients. It was seen that errors in the initial similarity values did not affect the flooding method very much.

*2.14. Anthropogenic Impacts on Ecological Connectivity* (**Saunders et al., 2015**)

This method crosses disciplinary and institutional boundaries, making it easier to identify trends in human impacts. Using graph theory, we reviewed research that has measured spatial functional connectedness in aquatic ecosystems. Forty two studies during the years 2000 to 2014 were found through the search. It evaluated whether each study had quantified the effects of four factors: overharvesting, climate change (warming temperatures, altered circulation or hydrology, sea-level rise, and ocean acidification), and human movements resulting in species introductions. The study's primary objective was to review how human actions impact functional connections in aquatic ecosystems. We conducted a quantitative assessment of research using metrics from graph theory to examine functional connectedness in aquatic ecosystems to bridge disciplinary and systemic gaps and establish a common approach and vocabulary. With the help of graph-based methods, the following aspects of the marine and freshwater ecosystems have been studied: (1) alteration of habitat (a decrease and deterioration, alteration of habitat associations, and creation of new habitats); (2) human-facilitated species movements; (3) overharvesting; and (4) changes in the climate and acidification of the oceans (warming water temperatures, altered circulation and hydrology, sea-level rise, and ocean acidification). This paper provides a platform for future testing of hypotheses of the impact of human-induced stressors on aquatic connection by summarising the application of graph theory in aquatic connectivity, synthesising across aquatic domains, and presenting the findings.

*2.15. Climate Forcing of Wetland Landscape Connectivity* (**McIntyre et al., 2014**)

In order to identify relationships between climatic factors and habitat connectivity for wildlife in present and prospective wetland landscapes and to understand how that capacity can alter as a consequence of climatic forcing, this study used wetlands as models in a graph-theory-based method. In order to investigate how the patterns and procedures that govern habitat connectivity change across landscapes, zones, and continents, we also give a case study of macrosystem ecology. On a macrosystems level, migratory bird connection among prairie wetlands is examined using graph theory. According to climate projections, there will likely be changes in the Great Plains' precipitation patterns, which will have an impact on the density of wetlands. Wetland networks show minimal connectivity and bird abundance under forecasted climatic scenarios.

*2.16. Sustainable Transportation Systems in Smart Cities* (**Nelson et al., 2018**)

The suggested remedy is centred around graph theory, a well-known scheduling-based issue that employs an implementation of software to automatically take in gathered data and generate suggestions based on it. A "one-fits-all" method that can be used as-is for all use cases, this method is very scalable and can be used in a general manner in the code. The Java code is based on Dijkstra's well-known shortest-path algorithm, which identifies the route across a collection of specified nodes in a network that requires the least amount of time, money, and effort. The system's accuracy is demonstrated by the number of shuttle records generated daily. The modelling approach is made more rigorous by the implementation of fines based on Elmwood Park delays, traffic, and weather conditions. The developed method can be used to improve traffic flow, minimise user annoyance, and lessen the carbon impact of transportation networks. The strategy used is based on the traditional scheduling-based problem of graph theory. Our graph theory approach pinpoints possible areas for traffic routes and transportation system improvement. Our methodology may be applied to a variety of situations to find transport networks with paths that could be improved, resulting in a reduction in CO2 emissions.

*2.17. Waste City Management System for Smart Cities Applications* (**Vu et al., 2017**)

The strategy of waste management in smart cities presented in this paper creates a clean urban environment at a low cost. In this method, data on trash volume is detected, measured, and transmitted via the Internet by a sensor model. Regression, classification, and graph theory are used to process the data that has been collected, as well as the serial number. A novel approach is now suggested to manage waste collection dynamically and effectively by forecasting waste status, identifying trash bin locations, and keeping track of waste production. The latter then suggests route optimisation to handle the garbage truck effectively. An algorithm that automatically creates functioning clusters determines the best garbage truck routes. Additionally, the total weight for every trash can was predicted and updated using logistic regression. They are going to be used to design the new, most efficient method for disposing of waste, which consumes less gasoline.

*2.18. Optimising* Air-Transportation Networks (**Solai 2021**)

An integral part of the communication system that underpins air travel is air traffic control. The method focuses on the use of the Kruskal method and Dijkstra's algorithm to search a network and find a Minimum Spanning Tree as a practical solution to the problem of determining the shortest path between two points. Additionally, a network model of the transportation issue was explicitly created for this case, and it is thoroughly examined to reduce shipping costs and locate the eccentricity along with Jordan's center. Dijkstra's solution basically looks for a connection between two vertices, whereas Kruskal's solution looks for a connection between 'n' vertices. Both flight planners and passengers want to minimise the fee, even though taking the longest route is expensive, but it will benefit airlines and passengers because there are more airports, and passengers will increase as to demand airport. Additionally, choosing the route with the fewest airports compromises safety because no airline can perform a quick manoeuvre to safety in the event of an emergency. In this situation, Jordan's central points can be useful for all the other airports in the structures for safety.

*2.19. IOT-based Air Pollution monitoring* (**P. Ferrer-Cid et al., 2022**)

Platforms for monitoring air pollution are crucial for both avoiding and lessening its consequences. Graph-based descriptions and analyses of air pollution monitoring networks are now possible because of recent developments in the field of graph signal processing. Reconstructing the measured signal in a graph using a portion of the sensors is one of the critical uses. Reconstructing the signal using data from nearby sensors is a crucial technique for preserving the integrity of network data. Examples include replacing missing data with information from correlated nearby nodes, creating virtual sensors, or correcting a drifting sensor with information from nearby sensors that are more precise. A graph signal reconstruction model is superimposed on a graph that was learned from the data in this article's proposed signal reconstruction framework for air pollution monitoring data. On actual data sets monitoring $O_3$, $NO_2$, and $PM_{10}$ levels of air pollution, various graph signal reconstruction techniques are contrasted. Both the capacity of the approaches to reconstruct a pollutant's signal and its computational cost are demonstrated. The findings highlight the superiority of kernel-based graph signal reconstruction techniques and the limitations of the techniques' ability to scale in an air pollution monitoring network with many inexpensive sensors. With the use of straightforward techniques, like splitting the network with a clustering algorithm, the framework's scalability can be increased.

*2.20. Analysing Criminal Networks by using Graph and Network Theory* (**Cavallaro et al. 2021**)

The study of social phenomena using social network analysis, which uses network and graph theory, has been proven to be extremely useful in fields like criminology. An actual case study is used in this chapter to offer an overview of the basic techniques and instruments that may be used to analyse criminal networks. We have taken information on the interactions between suspects in two Sicilian Mafia clans, starting from legal documents that are readily available, producing two weighted undirected graphs. Then, we looked into how these weights affected two crucial characteristics of the criminal network: weight distribution and shortest path length. We also discuss an investigation that seeks to create a synthetic network that mimics criminal activity. In order to achieve this, we used some of the most well-liked artificial network models, including Watts-Strogatz, Erd-Hos-Renyi, and Barab'asi-Albert, with some topological modifications, to compare the degree distributions of genuine criminal networks.

*2.21. Digital Twin: Doubling the Vision* (**Schrotter and H¨urzeler, 2020**)

The Digital Twin initiative in Zurich is a pioneering project designed to create a virtual representation of the city's physical landscape. Employing cutting-edge technologies like sensors, drones, and 3D modeling software, the project collects and amalgamates data concerning buildings, infrastructure, and public areas. This digital twin serves as a virtual replica, offering urban planners, architects, and engineers a platform to experiment with diverse urban design scenarios, forecast the repercussions of proposed alterations, and enhance the effectiveness of urban governance. Among its remarkable features, the digital twin has the capability to simulate urban climate scenarios and dissect the interactions between the built environment and various urban systems. Moreover, the platform elevates collaboration among both internal and external stakeholders, harnessing technologies such as virtual and augmented reality to provide immersive visualizations of proposed projects. Ultimately, the Zurich Digital Twin project is anticipated to elevate the city's sustainability, resilience, and overall quality of life. It is set to foster innovation and collaboration, setting a pioneering example for urban development initiatives worldwide.

## 3. Conclusion

The applications of graph theory algorithms in geospatial sciences are diverse and far-reaching. Graph theory gives geospatial professionals the tools to solve problems effectively, optimise resource allocation, and enhance decision-making in various domains, from modelling transportation networks to comprehending social relationships inside communities, planning delivery routes, and tracking environmental changes. The applications of graph theory in geospatial research are anticipated to rise further with the continuous development of technology and the increasing complexity of geographical data, making it a dynamic and essential topic for addressing the linked spatial concerns of our world. As geospatial data grows in volume and complexity, graph theory remains indispensable for making sense of spatial relationships and advancing the field of geospatial sciences.

This article underscores the critical role of graph theory algorithms in addressing real-world geospatial challenges, emphasising their practical significance and potential for future innovations in spatial analysis and management, including the geospatial digital twin concept.

**Acknowledgement**

This publication has been prepared as an output of the CGIAR Research Initiative on Digital Innovation. The authors would like to thank all funders who supported this research through their contributions to the CGIAR Trust Fund: https://www.cgiar.org/funders/

**Figure**

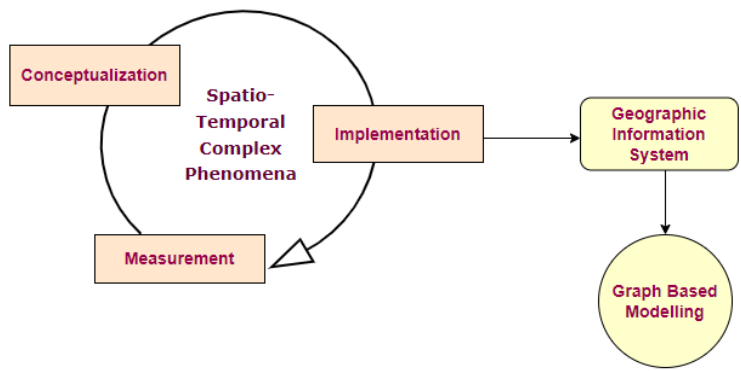

**Figure 1** *A conceptual framework representing the link between spatio-temporal phenomena, graph-based modelling and GIS* (simplified from Anderson et al., 2020)

**Tables**

**Table 1** Graph Terminologies

| Sl No | Item | Description |
|---|---|---|
| 1. | Graph | Graphs G = {V, E} are mathematical representations of networks that have vertices V and are linked by edges E. [Anderson et al., 2020]. |
| 2. | Vertices | Vertices are the fundamental units of the graph. They are also called nodes.[1] |
| 3. | Edges | Edges are drawn or used to connect two nodes of the graph. They are also called links. |
| 4. | Degree | The degree (or valency) of a vertex in a graph is determined by the total number of edges that are incident to it. |
| 5. | Path | A path is a track with no repeating vertices or edges. |

**Table 2** Types of Graphs

| Sl No | Type of Graph | Diagram | Description |
|---|---|---|---|
| 1. | Regular Graph | 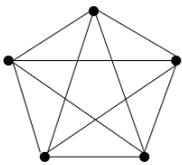 | A graph where every vertex has the same number of neighbours or the same degree or valency. |
| 2. | Random Graph | 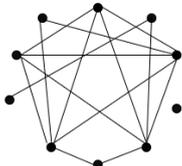 | Refers to probability distributions over graphs. They may be described simply by a probability distribution P(k) known as the Poisson distribution. |
| 3. | Undirected Graph | 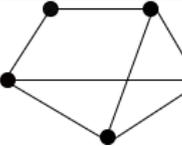 | A graph in which edges do not have any direction. |

---

[1] https://www.geeksforgeeks.org/mathematics-graph-theory-basics-set-1/

| 4. | Directed Graph | 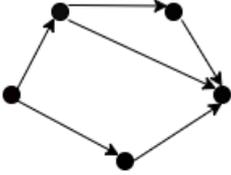 | A graph in which the edge has a direction. |
|---|---|---|---|
| 5. | Weight graph | 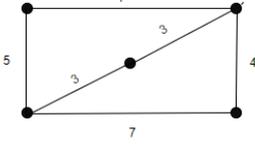 | A graph with a "weight" or "cost" at each edge. The weight of an edge can reflect distance, time, or anything else that represents the "connection" between the two nodes it connects. |
| 6. | Sparse Graph | 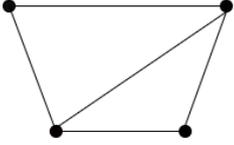 | A graph with few edges. |
| 7. | Dense Graph | 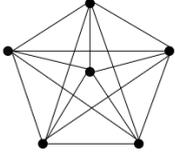 | A graph where the number of edges is about equal to the number of edges that can exist. |
| 8. | Tree | 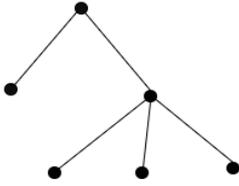 | A tree is an undirected graph with precisely one path connecting any two vertices. |

**Table 3** Graph Algorithms

| Sl No | Algorithm | Description |
|---|---|---|
| 1. | Greedy Algorithm | Greedy is an algorithmic paradigm that puts together a solution piece by piece, constantly choosing the subsequent element that offers the most obvious and immediate benefit. Greedy is, therefore, most appropriate to situations where selecting locally optimal results in a global solution.[2] |
| 2. | Dynamic Programming | Dynamic programming is a technique that divides problems into sub-problems and saves the solution for later use, eliminating the need to compute the result again.[3] |
| 3. | Dijkstra's Algorithm | Dijkstra's algorithm calculates the shortest distance between two places. It begins at the node we specify as the initial node in the graph and attempts to find the shortest route between it and every other node in the graph. It keeps track of the shortest path between each node and the source node at any given time. If a shorter path is discovered, the distance value is updated. The node is marked 'visited' and is added to the path if the shortest path is found. The technique is repeated until the route covers every node in the graph. The shortest route between the source node and all other nodes is obtained as a result. [Dijkstra et al., 1959]. |
| 4. | Prim's Algorithm | An illustration of a greedy algorithm that begins with an empty spanning tree is Prim's algorithm. The goal is to retain two sets of vertices while building a Minimum Spanning Tree (MST). Vertices that have already been included in |

---

[2] https://www.geeksforgeeks.org/greedy-algorithms/
[3] https://www.javatpoint.com/dynamic-programming

|   |   | the MST are in the first set, whereas those that haven't are in the second set. Consideration is given to all the edges between the two sets. Starting with source vertex, the tree grows in such a way that the vertex with the least weight from the edges is chosen and added to the tree. It selects the edge and then adds the opposite endpoint to the set containing MST. When the process is finished, the edges form a minimum spanning tree [Prim et al., 1957]. |
|---|---|---|
| 5. | Kruskal's Algorithm | Kruskal's algorithm is a more effective method for locating minimum-spanning trees on sparse graphs. All of the graph's edges are organised in ascending order using this approach. The edge with the lowest weight is chosen next, and a minimum spanning tree is built while keeping in mind that no cycles may happen when the edges are being added. Till the algorithm stops, this process is repeated. In order to prevent cycles from forming, a disjoint-set data structure is maintained. Two operations are available on the disjoint set: find and union, which determines whether a particular node belongs to a connected component. Problems involving route optimisation employ this algorithm. [Kruskal et al., 1956] |

**Table 4** Broad area of applications of graph theory in geospatial research

| Sl No | Thematic area | Scope for application of Graph theory |
|---|---|---|
| 1 | Urban planning | Graph algorithms can be used to analyse the connectivity of urban areas and to identify areas that are underserved by transportation or other infrastructure. This information can be used to improve the design of urban spaces and make them more livable. |
| 2 | Routing and navigation | Graph algorithms can be used to find the shortest paths between two points in a road network or to find the most efficient way to travel between a set of points. This is used in navigation systems, such as Google Maps and Waze. |
| 3 | Transportation planning | Graph algorithms can be used to plan new transportation routes or to optimise the use of existing transportation infrastructure. This is used in urban planning and in the design of transportation systems such as airports and seaports. |
| 4 | Disease spread modelling | Graph algorithms can be used to model the spread of diseases such as COVID-19. This can be done by creating a graph where the nodes represent people, and the edges represent connections between people. The spread of the disease can then be modelled by tracking the movement of people through the graph. This information can be used to identify high-risk areas and target interventions to prevent the spread of the disease. |
| 5 | Wireless network design | Graph algorithms can be used to design wireless networks, such as cellular networks and Wi-Fi networks. This can be done by creating a graph where the nodes represent wireless devices, and the edges represent connections between devices. The optimal placement of wireless access points can then be determined by finding the shortest paths between all pairs of devices. This information can be used to improve the performance of wireless networks by reducing interference and ensuring that all devices have a good signal. |
| 6 | Natural hazard analysis | Graph algorithms can be used to analyse the risk of natural hazards, such as floods and landslides. This can be done by creating a graph where the nodes represent features of the landscape, such as rivers, roads, and buildings, and the edges represent connections between features. The likelihood of a natural hazard occurring can then be modelled by tracking the movement of water or sediment through the graph. This information can be used to identify areas that are at risk of natural hazards and to take steps to mitigate the damage. |
| 7 | Environmental monitoring | Graph algorithms can be used to monitor the environment, such as tracking the movement of pollutants or wildlife. This can be done by creating a graph where the nodes represent features of the environment, such as rivers, lakes, and forests, and the edges represent connections between features. The movement of pollutants or wildlife can then be modelled by tracking the movement of water or animals through the graph. This information can be used to identify areas that are polluted or that are important for wildlife habitat. |